\documentclass[conference]{IEEEtran}
\IEEEoverridecommandlockouts
\usepackage{cite}
\usepackage{amsmath,amssymb,amsfonts}
\usepackage{algorithmic}
\usepackage{hhline}
\usepackage{multirow}
\usepackage{graphicx}
\usepackage{subfig}
\graphicspath{{Images/}}
\usepackage{textcomp}
\usepackage{xcolor}
\def\BibTeX{{\rm B\kern-.05em{\sc i\kern-.025em b}\kern-.08em
    T\kern-.1667em\lower.7ex\hbox{E}\kern-.125emX}}
\begin{document}

\title{SAHRTA: A Supervisory-Based Adaptive Human-Robot Teaming Architecture\\
\thanks{Funding Agency: NASA Cooperative Agreement No. NNX16AB24A and Department of Defense Contract Number W81XWH-17-C-0252 from the CDMRP Defense Medical Research and Development Program.
}
}

\author{\IEEEauthorblockN{Jamison Heard}
\IEEEauthorblockA{\textit{Dept. of Electrical Engineering} \\
\textit{Rochester Institute of Technology}\\
Rochester, NY \\
jrheee@rit.edu}
\and
\IEEEauthorblockN{Julian Fortune}
\IEEEauthorblockA{\textit{School of Electrical Engineering and Computer Science} \\
\textit{Oregon State University}\\
Corvallis, OR \\
fortunej@oregonstate.edu}
\and

\hspace{5.0cm}\IEEEauthorblockN{Julie A. Adams}
\IEEEauthorblockA{\hspace{5.0cm}\textit{Collaborative Robotics and Intelligent Systems Institute} \\
\hspace{5.0cm}\textit{Oregon State University}\\
\hspace{5.0cm}Corvallis, OR \\
\hspace{5.0cm}julie.a.adams@oregonstate.edu}

}

\textcopyright 2020 IEEE.  Personal use of this material is permitted.  Permission from IEEE must be obtained for all other uses, in any current or future media, including reprinting/republishing this material for advertising or promotional purposes, creating new collective works, for resale or redistribution to servers or lists, or reuse of any copyrighted component of this work in other works

\newpage
\maketitle
\thispagestyle{plain}
\pagestyle{plain}
\setcounter{page}{1}

\begin{abstract}

Supervisory-based human-robot teams are deployed in various dynamic and extreme environments (e.g., space exploration). Achieving high task performance in such environments is critical, as a mistake may lead to significant monetary loss or human injury. Task performance may be augmented by adapting the supervisory interface's interactions or autonomy levels based on the human supervisor's workload level, as workload is related to task performance. Typical adaptive systems rely solely on the human's overall or cognitive workload state to select what adaptation strategy to implement; however, overall workload encompasses many dimensions (i.e., cognitive, physical, visual, auditory, and speech) called workload components. Selecting an appropriate adaptation strategy based on a complete human workload state (rather than a single workload dimension) may allow for more impactful adaptations that ensure high task performance. A Supervisory-Based Adaptive Human-Robot Teaming Architecture (SAHRTA) that selects an appropriate level of autonomy or system interaction based on a complete real-time multi-dimensional workload estimate and predicted future task performance is introduced. SAHRTA was shown to improve overall task performance in a physically expanded version of the NASA Multi-Attribute Task Battery. 

\end{abstract}


\section{Introduction}

Human supervisors perform mission critical tasks in a wide range of dynamic environments, such as space exploration. It is imperative that high task performance is achieved, as a mistake may lead to mission failure. The human supervisors may experience erratic workload levels \cite{Wickens2004,Sim2008}, where performance tends to decline when workload is too high (overload) or too low (underload) \cite{Wickens2004}. These undesirable workload states may be mitigated by targeting system interactions or autonomy levels towards the human's workload state; thus, augmenting task performance. The supervisory-based adaptive human-robot teaming architecture (SAHRTA) that augments task performance by targeting adaptations towards a continuous and complete estimate of the human's workload state is introduced. 

Typical supervisory-based adaptive human-machine or human-robot interfaces select an appropriate adaptation strategy (e.g., autonomy levels) based on a discrete cognitive workload measurement (e.g., \cite{Kaber2004,Schwarz2018}). Relying solely on cognitive workload limits the adaptive system's ability to reason how an adaptation affects the human, as overall workload has many contributors (i.e., cognitive, physical, visual, auditory, and speech \cite{Mitchell2000}) and conflicts may exist between the contributors. For example, a human is overloaded due to a resource conflict between the auditory and speech workload channels (e.g., verbally communicating mission critical information, while audible alarms sound). A system relying solely on cognitive workload may reason that the human is overloaded and that an increase in the system's level of autonomy will reduce the human's workload level. However, such an adaptation may not actually reduce the human's speech or auditory workload channels, which are the primary contributors to the human's overloaded state. A system that is cognizant of the human's complete workload state can identify that a resource conflict exists and modify the interaction modality of the audible alarm (e.g., visual or tactile modality) in order to resolve the conflict.

Another limitation of the state-of-the-art adaptive systems is that the systems rely on a discrete representation of workload (i.e., underload, normal load, or overload). This discrete representation may be beneficial in determining if the system's autonomy level needs to change, but does not provide information regarding the level of autonomy change needed. For example, increasing the autonomy level from full manual to full autonomy will likely mitigate a human's overloaded workload state, but may inadvertently place the human in an underload state. An adaptive system may change the autonomy level incrementally in order to circumvent this problem. Changing the autonomy level incrementally requires time in order to mitigate undesired workload states appropriately and requires constantly assessing if another incremental autonomy level change is required. A continuous workload representation will allow a system to better determine how much the autonomy level needs to change; thus, potentially mitigating an undesired workload state quicker.

The research contribution focuses on improving supervisory human-machine teams by developing SAHRTA, which relies on a real-time multi-factoral continuous workload assessment algorithm in order to target intelligent adaptations (i.e., adapting autonomy levels or interaction modalities) and improve task performance. A pilot-study demonstrated the system's effectiveness. This paper is organized as follows: Section II describes work related to adaptive systems, while Section III introduces the task environment in which the adaptive system was deployed. Section IV presents the adaptive system's architecture, followed by the results being presented in Section V. Section VI discusses the results and conclusions.

\section{Related Work}

Developing an adaptive human-robot teaming system requires an architectures to facilitate when and how an adaption occurs. Typical adaptive system architectures prescribe to the ``perceive, select, act'' cycle \cite{Wickens1992}, where the system perceives some state variable, selects an action to perform based on the state variable, and acts by implementing the action. 

% The system must ``perceive'' some mechanism in order to invoke or disengage adaptions. These mechanisms can pertain to system, world, task, spatio-temporal, or human states \cite{Feigh2012}. The system state encompasses known system knowledge, such as current or predicted operation modes. The world state uses environmental measures, such as ambient light, to gain understanding of the surrounding environment. The task state corresponds to the current allocated task set, but can also be abstracted to mission variables (e.g., mission plan or human intent). Spatio-temporal states incorporate location information and time information, while human states may be determined by workload or engagement. 

% The ``select'' state may choose the adaptation types, which can be categorized as function allocation, task scheduling, interaction, and content \cite{Feigh2012}. Function allocation determines what tasks are allocated to which agents (human or system), while task scheduling can change when tasks are performed or the tasks' priority levels. A system may dynamically change its interactions by changing the interaction modality (e.g., auditory or visual) of a stimulus. The system can also vary the amount of content available to the human, such as reducing visual clutter in a high workload condition.  Finally, the system ``acts'' or adapts once it selects an adaptation type(s), either by changing the system automation level or the system's interactions with the human.

Adaptive systems focus primarily on adapting autonomy levels (i.e., adaptive automation). Adaptive automation frameworks typically rely on the human's cognitive workload state to allocate control to the system or human  \cite{Kaber2004, Sheridan2011}. Higher levels of automation may elicit the underload state \cite{Lin2018}, while lower levels of automation may elicit the overload state. Thus, adaptive automation may use workload estimates to prevent underload and overload states from occurring, or mitigating them when they occur. An open research question is how often to switch levels of autonomy, as frequent switching may create a ``yo-yo effect'' that causes increased workload \cite{Abbass2014}. Switching infrequently may not improve task performance.

Adaptive systems are not limited to adaptive automation frameworks. Task difficulty may be varied, rather than system autonomy. Bian et al. \cite{Bian2019} varied the difficulty of a driving task based on human engagement and measured performance. The participants were more engaged when the task adapted to their engagement level, than when no adaptation occurred, which illustrates that a desired engagement level may be obtained by varying task difficulty. Walter et al. \cite{Walter2017} manipulated task difficulty based on EEG-based cognitive workload measurements. The participants with the adaptive task difficulty had significant learning effects, demonstrating that manipulating task difficulty based on workload measurements is feasible.

Most adaptive systems have focused on a single human state construct to determine adaptations, but a multi-dimensional adaptation scheme has been theorized \cite{Schwarz2018, Fuchs2017}. The ``Real-Time Assessment of Multidimensional User State'' system was intended to assess cognitive workload, fatigue, and attentional focus in order to adapt a system's interactions intelligently, but only recognized high workload or high fatigue. This multi-dimensional human state assessment was fed into an adaptation decision framework in order to determine how an adaptation occurs. The proof-of-concept system automated tasks to reduce cognitive workload, visually highlighted high priority tasks to correct attentional focus, and used an auditory modality for alarms if the operator was fatigued. Although the system seems promising, no performance data was presented; thus, the ability to improve performance was not provided. Further, it is unclear if the three participants actually experienced high fatigue levels during the 45-minute task, or if the system can adapt to the underload workload state.

Although the current adaptive systems seem promising; none of theses systems are capable of adapting interactions and autonomy levels based on a complete estimate of the human's workload state (i.e., cognitive, physical, visual, auditory, and speech), while accounting for underload and overload conditions. SAHRTA was developed to overcome these limitations.

\section{Task Environment}
\label{sec:adaptive_teaming}
 
\begin{figure*}[!h]
    \subfloat[Tracking]{ \includegraphics[width=.45\textwidth, height=3.0cm, trim = {0cm 0cm 0cm 0cm}, clip]{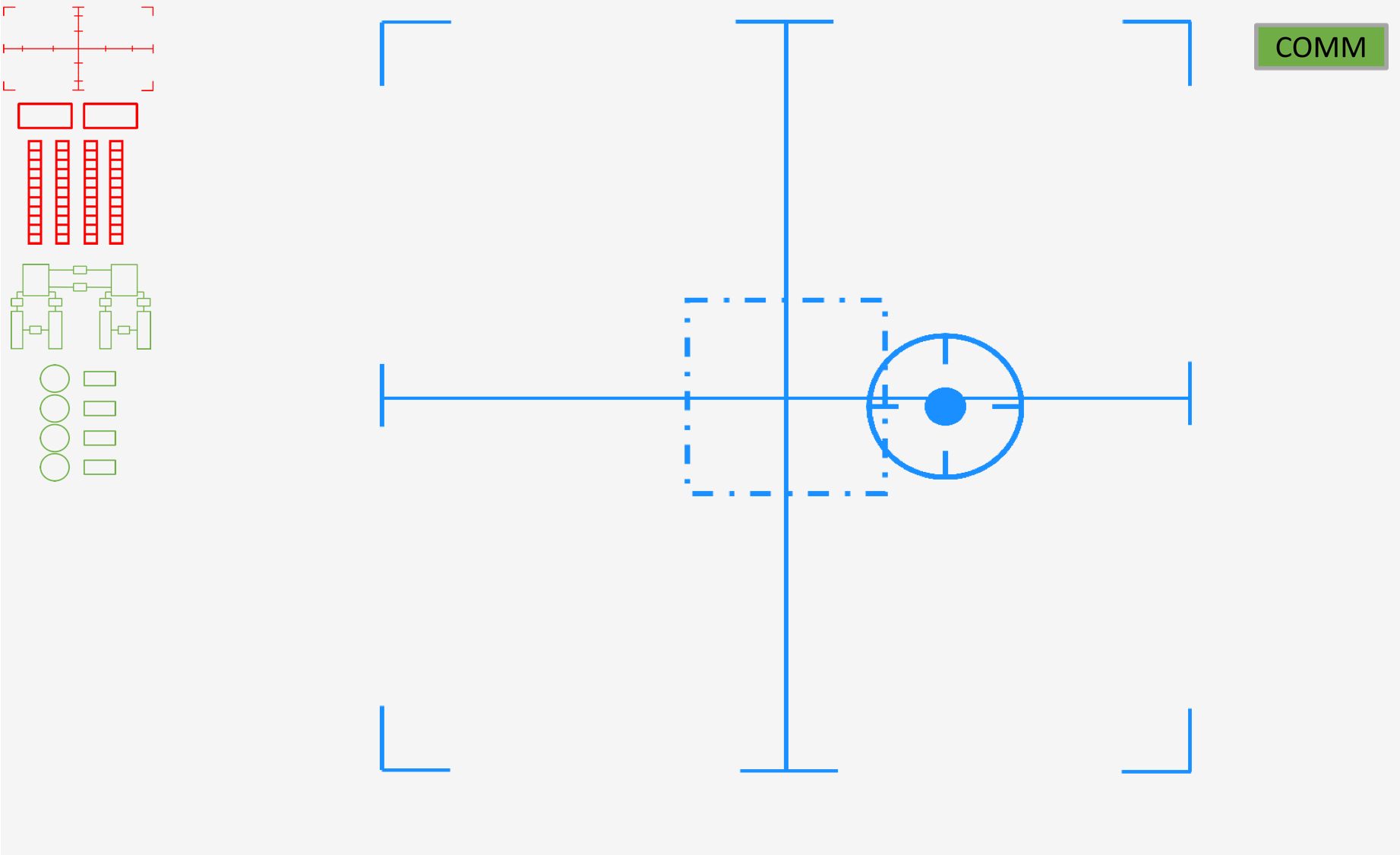}} \hspace{1.0cm}
    \subfloat[System Monitoring]{\includegraphics[width=.45\textwidth, height=3.0cm, trim = {0cm 0cm 0cm 0cm}, clip]{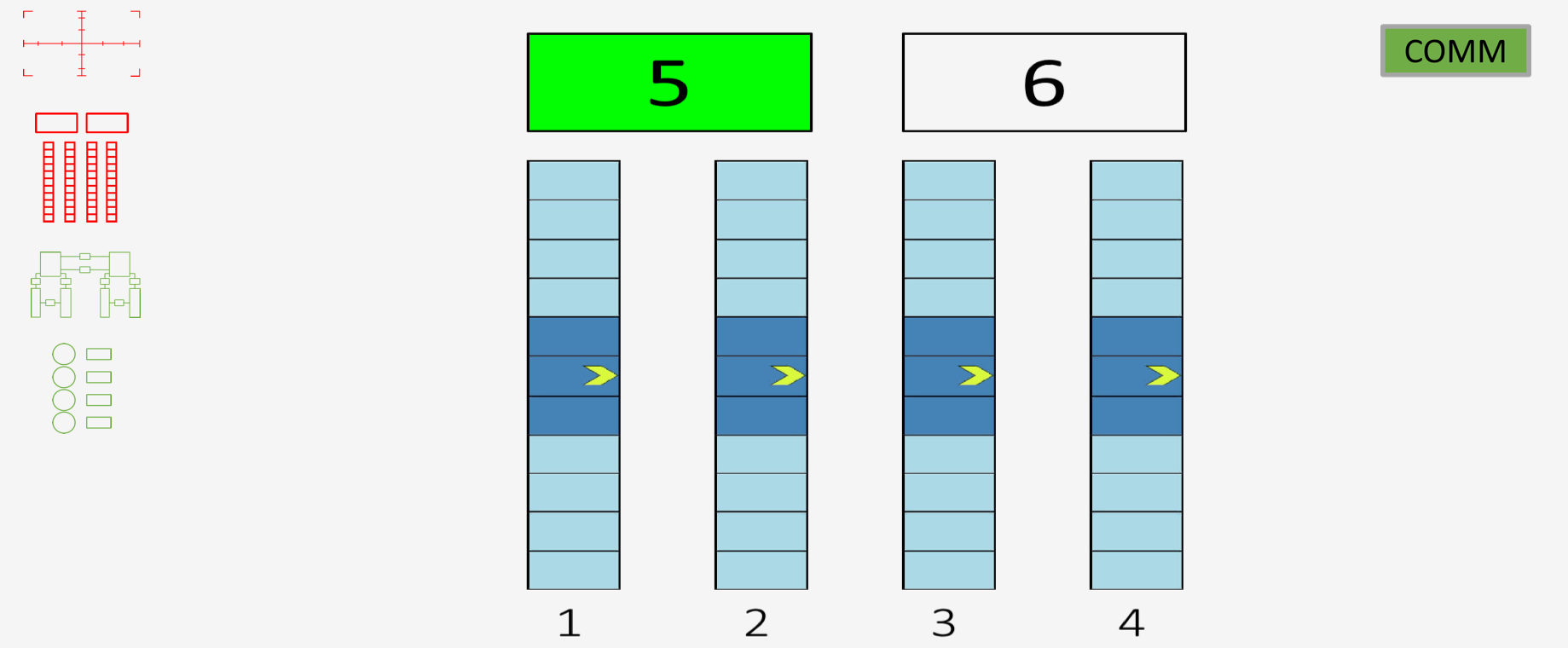}}
    \newline
    \subfloat[Resource Management]{\includegraphics[width=.45\textwidth, height=3.0cm, trim = {0cm 0cm 0cm 0cm}, clip]{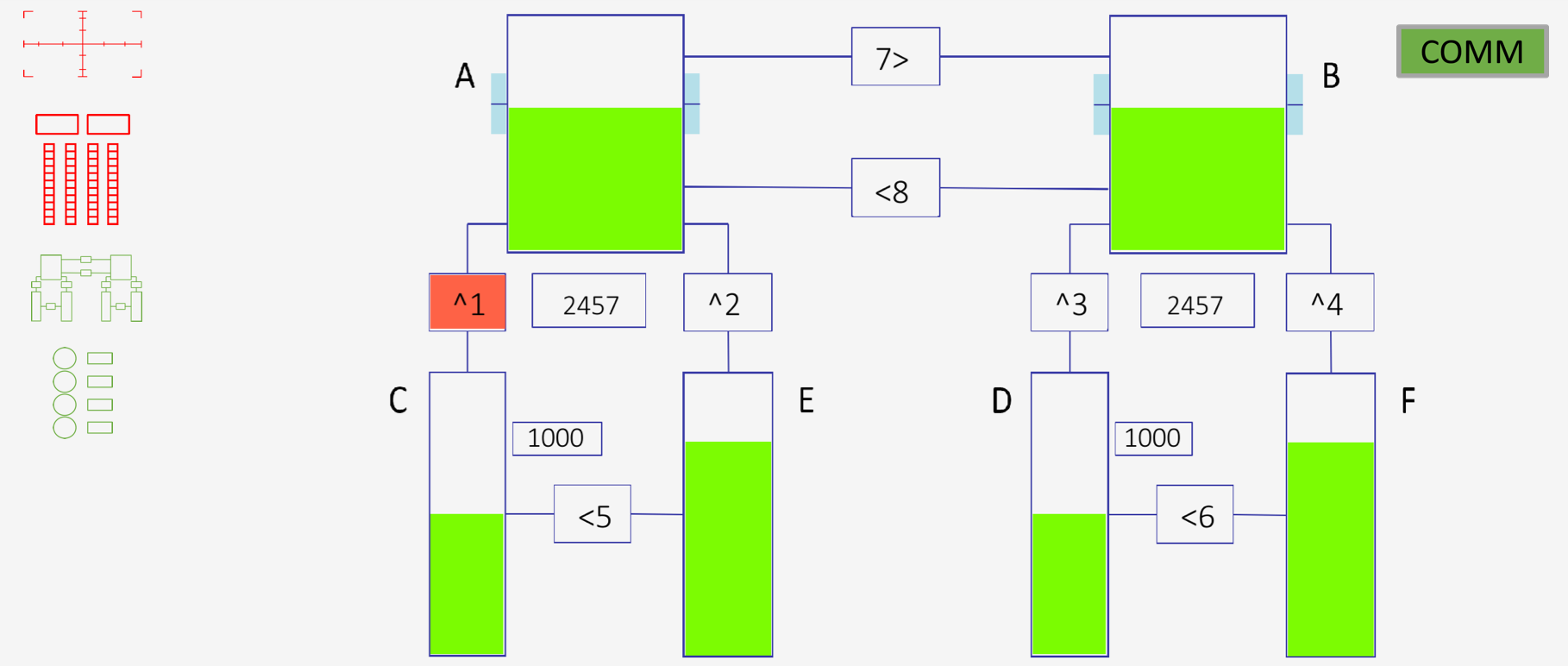}} \hspace{1.0cm}
    \subfloat[Communications]{\includegraphics[width=.45\textwidth, height=3.0cm, trim = {0cm 0cm 0cm 0cm}, clip]{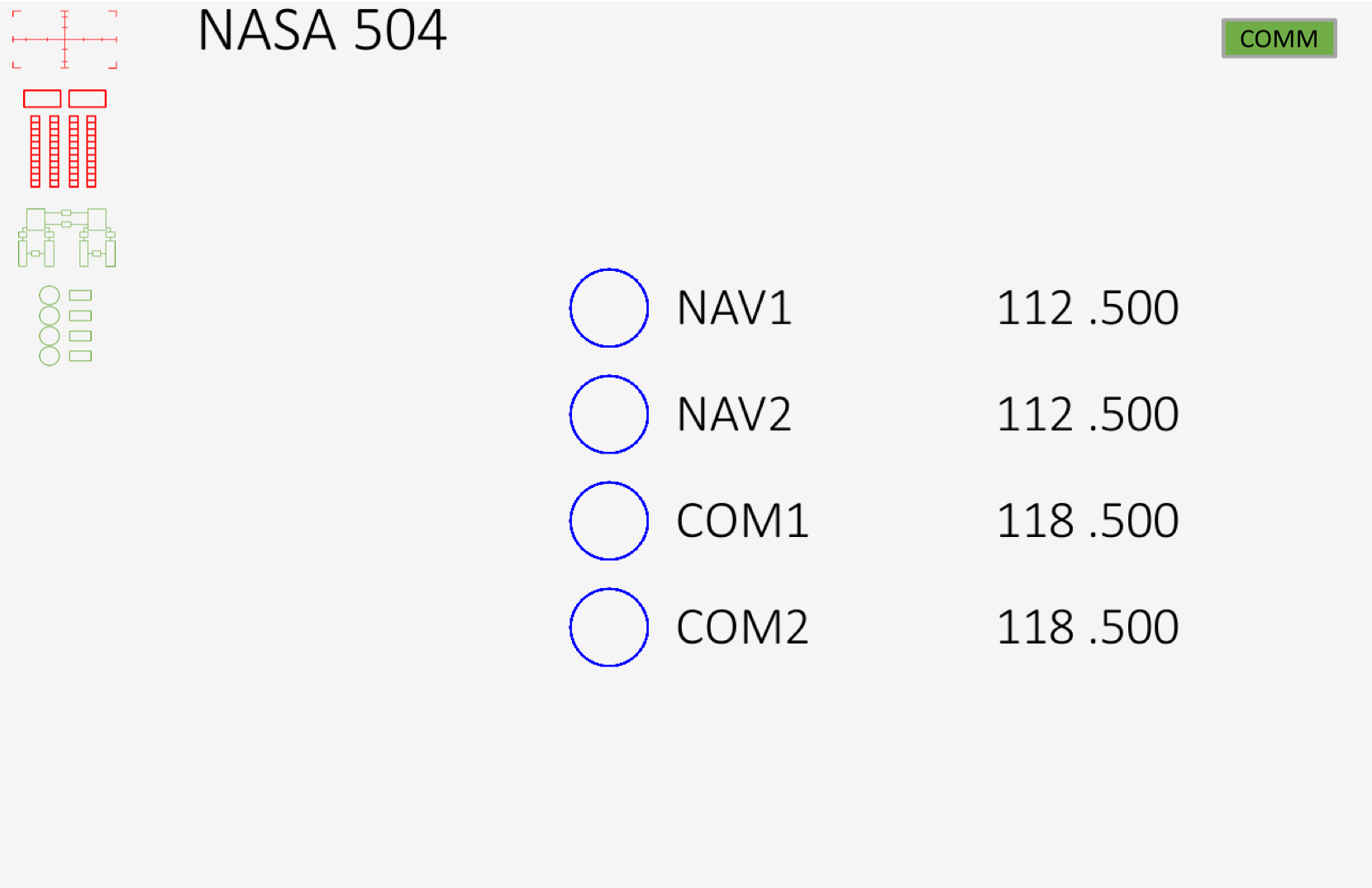}}
    \caption{The NASA MATB-II Tasks}
    \label{fig:adapted_matb}
\end{figure*}

SAHRTA was developed to be used in a wide-range of human-machine teaming domains, but the presented study focused on applying the architecture to a physically expanded version of the NASA Multi-Attribute Task Battery (NASA MATB-II) \cite{Comstock1992}. The original NASA MATB-II simulated supervising a remotely piloted aircraft and consisted of four concurrent tasks: tracking, system monitoring, resource management, and communications. Workload was manipulated (underload (UL), normal load (NL), or overload (OL)) by changing various parameters of each task in order to determine the adaptive teaming system's effectiveness.

The tracking task, depicted in Figure \ref{fig:adapted_matb} (a), required participants to keep the circle with a blue dot in the middle of the cross-hairs using a joystick. Workload was manipulated by setting the tracking mode: automatic (low) or manual (high). The underload condition used the automatic mode, with no input from the human, while the overload condition required the manual mode, or full human control. The normal workload condition switched between manual and automatic modes approximately every 2.5 minutes.

The system monitoring task required monitoring two colored buttons and four gauges, shown in Figure \ref{fig:adapted_matb} (b). If the green button turned grey (off) or the other button turned red (on), the value was considered out of range and required resetting by selecting the button. The four gauges had a randomly moving indicator, up and down, that typically remained in the middle. The participants were to click on a gauge if it was out of range (i.e., the indicator was too high or too low). The underload condition had one out of range instance per minute, overload had twenty out of range instances per minute, and normal load had five instances per minute. 

The resource management task included six fuel tanks (A-F) and eight fuel pumps (1-8), shown in Figure \ref{fig:adapted_matb} (c). The arrow by the fuel pump's number indicated the direction fuel was pumped. Participants were to maintain the fuel levels of Tanks A and B by turning the fuel pumps on or off. Fuel Tanks C and D had finite fuel levels, while Tanks E and F had an infinite fuel supply. A pump turned red when it failed. Zero pumps failed during the underload condition, while two or more pumps failed per minute during the overload condition. The normal load condition switched from zero pumps failing to one or two pumps failing every minute. 

The communications task required listening to air-traffic control requests for radio changes. A communication request was ``NASA 504, please change your COM 1 radio to frequency 127.550.'' The original MATB communications task required no speech, but a required verbal response was added. A response may be ``This is NASA 504 tuning my COM 1 radio to frequency 127.550.'' Participants were to change the specified radio to the specified frequency by selecting the desired radio and using arrows to change the radio's frequency, as depicted in Figure \ref{fig:adapted_matb} (d). Communications not directed to the participants' aircraft, as indicated by the call sign, were to be ignored. The  The underload condition contained $\leq$ 2 requests per minute, the overload contained $\geq$ 8 per minute, and normal load contained two to eight requests per minute.

 \begin{figure}[!h]
     \centering
     \includegraphics[width=.5\textwidth, height=6.0cm, trim = {5cm 3cm 0cm 2.5cm}, clip]{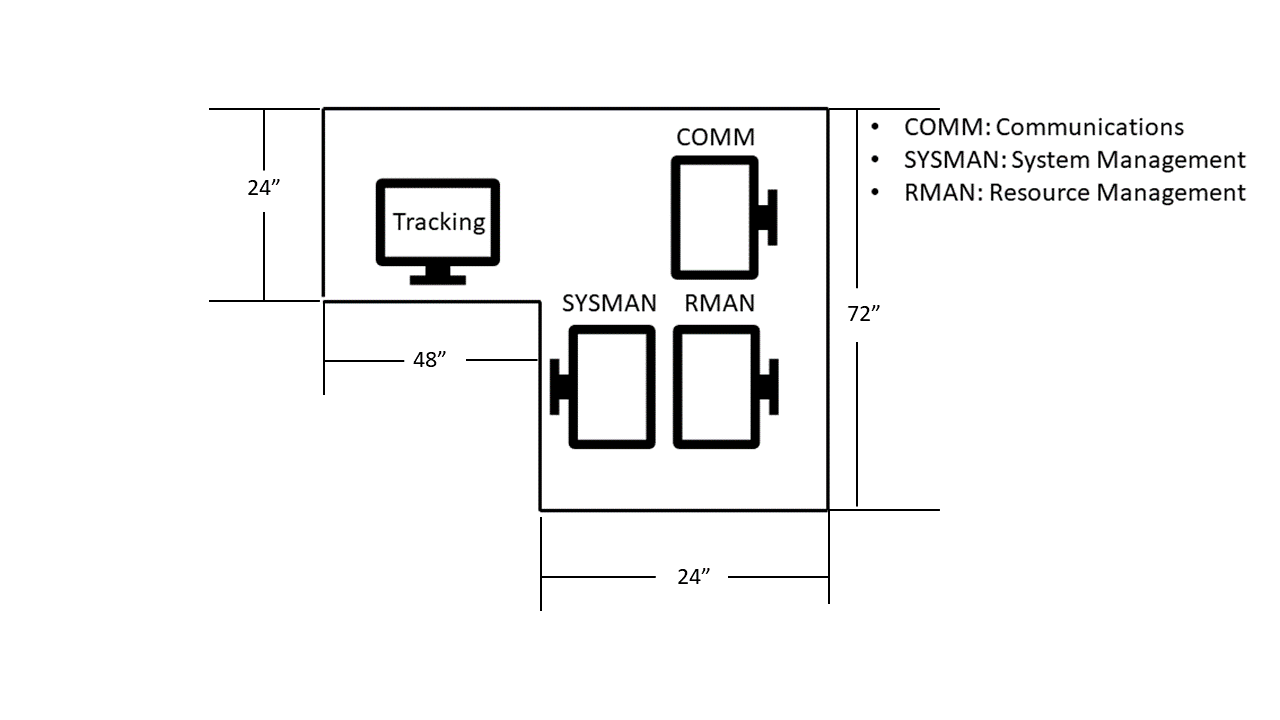}
     \caption{Physical Layout of the Adapted NASA MATB-II}
     \label{fig:layout}
     \vspace{-.5cm}
\end{figure}

\begin{figure*}[!h]
		\centering
		\includegraphics[width=0.75\linewidth, height = 5.5cm]{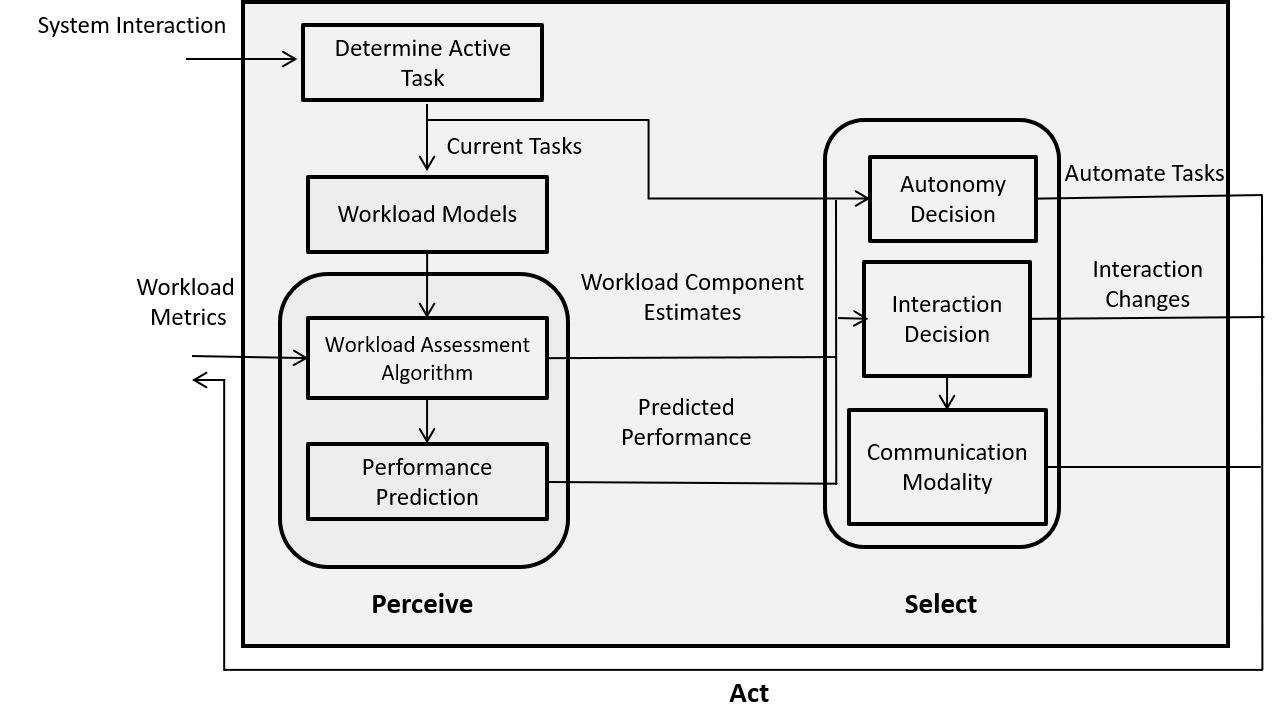}
		\caption{The SAHRTA Architecture.}
		\label{fig:adaptive_system}
		\vspace{-.5cm}
\end{figure*}

The original NASA MATB-II had all tasks on a single screen; thus, participants remained stationary. There are supervisory-based environments that require movement throughout the environment (e..g, a nuclear power-plant). Thus, the NASA MATB-II was adapted to require movement throughout the task environment by physically separating each task. This physical layout is depicted in Figure \ref{fig:layout}. Each task had a computer monitor dedicated to the particular task, where the computer monitors were stationed such that the participant was unable to visually see more than two tasks simultaneously. This visual hindrance ensured that participants walked around the task environment. The required equipment (e.g., joystik or a keyboard) to complete each task was placed in front of the respective monitor. The table surfaces were approximately 4 ft. from the floor. Participants were free to tilt the computer monitors in order to accommodate height differences.

The physically expanded version of the NASA MATB-II was coded using Python and PyGame in order to have more control over the task environment. The same task parameters (e.g., tank fuel rates) from the original NASA MATB-II were reimplemented. Information regarding the fuel pumps' rates and task scheduling was omitted in order to reduce the visual screen clutter. Each monitor screen is depicted in Figure \ref{fig:adapted_matb}. There is no current method to combine each task performance measure into an overall measure. Each performance measure was mapped to a value from 0 to 1 in order to permit combining the measures, where 1 represents optimal performance. The tracking task performance (i.e., root-mean square error (RMSE) between the center of the crosshairs and the object) was normalized based on participant data. The system monitoring task's and communications task's performance were measured using two metrics: reaction time and success rate. Reaction time was the time delta in seconds from when an out-of-range light, gauge, or communications request went out of range to when the participant corrected the out-of-range instance. Success rate represents the number of out-of-range instances corrected divided by the total number of instances. Reaction time was normalized, while success rate was already within range. A value of 1 was assigned if the resource management task's fuel levels were within 2,000 and 3,000 units, while the tank levels were normalized outside of that range. The overall performance measure was the uniform average of all active tasks' performance measures. If the resource management and system monitoring tasks were the only active tasks, then the overall task performance was the average of those two tasks' performance measures.

Normalizing the performance metrics and using an uniform average calculation may not be the optimal solution to generating an overall performance score. Normalizing performance data does not penalize time dependent measures. For example, the fuel tank levels can only rise so quickly; thus, fuel levels much smaller than 2,000 units need to be penalized more than fuel levels close to 2,000 units. However, developing appropriate time penalizations is not trivial and tangential to assessing SAHRTA's ability to improve performance. Additionally, using an uniform average to calculate overall task performance does not account for task priority levels. The participants were not given any priortiziations; thus, the use of an uniform average.

\section{SAHRTA}
 
SAHRTA was composed of three stages: \textit{Perceive}, \textit{Select}, and \textit{Act}. An overview of the architecture is provided in Figure \ref{fig:adaptive_system}. The \textit{Perceive} stage consisted of a workload assessment algorithm and a performance prediction model. The algorithm provided real-time estimates of overall workload and its contributing components \cite{Heard2019HFES, Heard2019thri, Heard2019jcedm}. The algorithm extracted features (i.e., mean, variance, average gradient, and slope) from thirty seconds epochs for each objective workload metric (i.e., heart-rate, heart-rate variability, respiration-rate, posture magnitude, noise level, speech-rate, speech intensity, and pitch). These extracted features were inputs to a corresponding neural network, which estimated a workload component every five seconds. There was a separate neural network for each workload component, where the component estimates are uniformly aggregated to estimate overall workload. This overall workload estimate was mapped to a state (i.e., underload, normal load, or overload) using thresholds. % (e.g., a value $\geq$35 corresponded to the overload state)

Contextual features calculated from workload models developed using IMPRINT Pro \cite{Archer2005} were required for accurate workload estimates. These features required knowing the participant's current task; however, knowing this task in dynamic domains is not trivial. The presented system version assumed that the participant's current active task was always known. The interface tracked the participant's last input (e.g., moving the joystik or a keystroke), which corresponded to the participant's current task. The participants often completed more than one task simultaneously; thus, the closest task to the participant's last input was included in the current task set. For example, if the participant moved the joystik, then the current task set consisted of the tracking and system monitoring tasks. %Likewise, if a participant tuned a radio, then the task set consisted of the resource management and communication tasks.  
 
Future task performance was predicted by a performance prediction model, which relied on a long short-term memory neural network architecture \cite{Hochreiter1997}. Long short-term memory networks use the previous time-step information to predict future time-steps in a sequential data series. The developed model used the last three workload estimates (i.e., overall workload and each workload component) as inputs. The performance model consisted of three long short-term memory layers, each with 256 neurons. Each neuron in the long short-term memory layers had an 80\% chance to dropout during training, which meant that the neuron may be excluded from training activation and weight updates \cite{Hochreiter1997}. There was a 256 neuron fully connected layer with a rectified linear unit activation function after the three long short-term memory layers. The model's output regression layer predicted overall task performance for one minute into the future. The ADAM optimizer \cite{Kingma2015} with a mean-squared error loss function was used to train the performance prediction model.
 
The \textit{Select} stage identified if a task needed to be automated or how an interaction occurred using the knowledge of the human's current task set, the workload estimates, and predicted performance. If the human's predicted performance fell below a threshold value (0.70), or when the last three overall workload estimates were in the overload state, then all inactive tasks (as determined by the interface) were transitioned to automation mode. Three workload estimates was chosen to ensure that the system did not thrash cyclically, causing the automation to turn on and off for each workload estimate. If the last three overall estimates were considered in the underload state, or the human's predicted performance was above a threshold level (0.85), then all tasks transitioned out of automation mode. The threshold levels were chosen based on the overall performance values previously collected \cite{Heard2019jcedm}.

The adaptive teaming system architecture determined how a system interaction occurred, once an interaction was expected to occur. These interactions occurred when the tracking task switched modes (e.g., manual to automation), when the system monitoring task's light or gauge went out of range, or when the resource management task's fuel levels went out of range. The adaptive system selected a communication modality (i.e., visual or auditory) based on potential conflicts in the workload channels. A visual modality was used if the participant's visual workload channel was not overloaded, meaning that the participant had sufficient resources to parse the interaction's visual information. An auditory modality was used if the human's speech and auditory workload channels were not loaded, as an auditory stimulus may distract the participant if they were speaking, or if there was substantial environmental noise. The interaction's auditory stimulus was postponed for 5 seconds, if the participant's auditory or speech channels were loaded. If after 5 seconds the workload channels were still loaded, then the interaction used a visual stimuli only.  

Interactions pertain to how the system conveyed information and how the participant interacted with the system (e.g., clicking a mouse). The \textit{Select} stage changed the communication task's interaction modality, depending on the participant's available resources. The participants were able to speak in order to change the communications task's radios, instead of using a physical modality (i.e., using the mouse). The participants were told that the system used speech recognition to determine what radio and frequency they were saying, but the system detected that the participant was speaking and assumed that they said the correct radio/frequency. 

Icons, on the left side of each computer screen (Figure \ref{fig:adapted_matb}), were used to communicate each task state (e.g., the task was being automated or not). An icon was green if the task was in automation mode, red if the task was out of range (e.g., a light went out of range), or grey if the participant was to determine if a task was out of range. The icons were greyed when the visual channel was determined to be overloaded and the corresponding task was not being automated in order to reduce visual workload. There was an interaction icon that appeared on the right side of each computer screen (Figure \ref{fig:adapted_matb}), that represented when the participant was able to interact with the communications task via a speech modality.

\section{Experimental Design}

The pilot study used a mixed-experimental design, with workload and adaptation condition as the independent variables. The workload conditions were underload (UL), normal load (NL), and overload (OL). Seven consecutive 7.5 minute workload conditions (OL-UL-OL-NL-UL-NL-OL) composed the 52.5 minute trial. This ordering was chosen such that each participant experienced each workload transition (e.g., UL to OL) in order to emulate real-world conditions \cite{Heard2019jcedm}. The dependent variables consisted of the workload algorithm's estimates, performance, and subjective metrics. 

A total of eighteen participants completed the trial in either a \textit{No Adaptation} condition or in one of three adaptation conditions (\textit{Autonomy}, \textit{Interaction}, and \textit{Both}), allowing for a between-subjects analysis. SAHRTA was not used during the \textit{No Adaptation} condition, while SAHRTA automated tasks during the \textit{Autonomy} condition, but did not adapt interactions. Likewise, SAHRTA adapted interactions during the \textit{Interaction} condition, but did not automate tasks. The \textit{Both} condition used SAHRTA to automate tasks and adapt interaction modalities. 

Ten participants completed the \textit{No adaptation} condition. The remaining eight participants completed two adaptive conditions: \textit{Both} and either \textit{Autonomy} or \textit{Interaction}, where one condition was completed in the trial's first half and the other was completed in the second half. The adaption conditions were counterbalanced among the eight participants resulting in four ordering pairs (e.g., \textit{Interaction} and \textit{Both}).

\subsection{Procedure}
 
The participants completed a consent form and a demographic questionnaire upon arrival, after which participants were fitted with a BioPac Bioharness\textsuperscript{TM}, a Schure Microphone, and two  Myo  devices. The Myos collected electromyography and acceleration data from the participant's forearms. A 15-minute training session occurred prior to commencing the 52.5-minute trial. The NASA Task-Load Index \cite{Hart1988} and a post-session questionnaire were completed upon trial completion. In-situ workload ratings \cite{Harriott2013} were collected 7 minutes into the trial and every 7.5 minutes after the initial collection.
 
The eighteen participants (8 female and 10 male) had a mean age of 24.9 (St. Dev. = 1.72), seven of which were graduate students in the Robotics program at Oregon State University. Ten participants held an undergraduate degree and eight participants held a Master's degree. Participants rated their video game skill level on a Likert Scale (1-little to 9-expert) with an average of 4.90 (St. Dev. = 2.42). Seven participants played video games at most 3 hours a week. Seven participants drank no caffeine the day of the experiment, while three participants drank 16 oz or less.
 
The participants slept on average 6.75 (St. Dev. = 1.51) hours the night before the experiment and on average 7.80 (St. Dev. = 1.03) hours two nights prior. The participants rated their stress and fatigue levels on a Likert scale (1-little to 9-extreme) with an average stress level of 2.7 (St. Dev. = 1.06) and average fatigue level of 3.3 (St. Dev. = 1.80). %No participant was determined to be an outlier based on their performance and subjective data. %Thus, the participant demographics did not impact or marginally impacted the results.
 
% \subsection{Metrics}
 
% An overview of the metrics collected is provided in Table \ref{tab:adaptive_metrics}.

% \begin{table}[!h]
%     \centering
%     \caption{The Objective and Subjective Metrics for the Real-Time Evaluation.}
%     \label{tab:adaptive_metrics}
%     \begin{tabular}{c|l} \hline
%          \textbf{Metric Type} & \textbf{Metric} \\ \hline
%         \multirow{9}{*}{Algorithm} & Heart-Rate \\
%          & Heart-Rate Variability \\
%          & Respiration-Rate \\
%          & Posture \\
%          & Noise Level \\
%          & Speech-Rate \\
%          & Pitch \\
%          & Voice Intensity \\ \hline
%         \multirow{8}{*}{Other Objective} & Body Activity \\
%          & Arm Acceleration \\
%          & Forearm Electromyography \\ 
%          & Tracking Task: Tracking Error \\
%          & System Monitoring Task: Reaction Time \\
%          & System Monitoring Task: Failure Rate \\
%          & Resource Management Task: Time-in-Range \\
%          & Communications Task: Reaction Time \\ \hline
%          \multirow{2}{*}{Subjective} & In-Situ Workload Ratings \\
%          & NASA-TLX \\ \hline
%     \end{tabular}
    
% \end{table}

\subsection{Hypotheses}

Two hypotheses were formulated in order to determine if the adaptive system is effective in augmenting task performance. It was expected that the adaptive system will have a significant effect on human workload. Specifically, the adaptive system can neutralize workload by lowering workload in the overload condition and increasing workload in the underload condition. Hypothesis $\mathbf{H^{A}_1}$ predicted that the workload assessment algorithm's estimates will differ between the \textit{No Adaptation} and \textit{Adaptation} conditions, with lower workload experienced in the overload condition and higher workload experienced in the underload condition when using SAHRTA. Neutralizing workload may affect task performance; thus, Hypothesis $\mathbf{H^{A}_2}$ predicted that higher performance will be achieved when using SAHRTA for each NASA MATB-II task.

\section{Adaptive System Results}
 
%The adaptive teaming system was analyzed from four perspectives: between evaluation, adaptive autonomy, interaction selection, and within-subjects. The between evaluation analysis compares workload estimates, performance, and subjective workload ratings between ten participants of the real-time evaluation (\textit{No Adaptation}) and eight participants from the adaptive system evaluation (\textit{Adaptation}). The adaptive autonomy and interaction selection analyses investigate the impact of adapting system autonomy and interactions, respectively. The within-subjects analysis compares the same metrics as the between evaluation analysis for two participants, whom completed both the real-time and adaptive evaluations. Statistical analysis was not conducted on the data due to insufficient power to find meaningful differences across the participants.

SAHRTA was designed to improve task performance by managing a human's workload state; thus, it was expected that participants will experience different workload levels when using SAHRTA (\textit{Adaptation}) and not using the system (\textit{No Adaptation}). The workload assessment algorithm's estimates by workload and adaptation condition are provided in Table \ref{tab:wl_estimates_adaptive}. The workload estimate's theoretical ranges are as follows: auditory (0-4), cognitive (0-22), physical (0-12), speech (0-4), and overall (0-62). The \textit{Adaptation} condition's results are an aggregate of the \textit{Both}, \textit{Autonomy}, and \textit{Interaction} conditions in order to see an overall workload affect when SAHRTA was being used vs. when not being used. The workload estimates, other than physical workload, were lower when using SAHRTA for the overload condition. Similar physical workload estimates occurred in the overload condition, which is attributed to the participants primarily remaining stationary during the workload condition. The participants experienced higher overall workload during the underload condition, due to the adaptive system allocating the tracking task to underloaded participants. Auditory workload was also higher using the adaptive system, as an auditory stimulus was used to alert participants of the out of range tasks. The participants experienced similar workload levels when using SAHRTA vs. not using SAHRTA during the normal load condition.

\begin{table}[!h]
\centering
\caption{\label{tab:wl_estimates_adaptive} Algorithm Estimated Workload by Condition and Adaptation Type: No Adaptation (None) vs. Adaptation.}

\begin{tabular}{c c c c c} \hline

\textbf{Workload} & \textbf{Adaptation} & \textbf{UL} & \textbf{NL} & \textbf{OL} \\ \hline

\multirow{2}{*}{Auditory} & None & 0.71 (1.41) & 2.33 (1.37) & 3.09 (1.02)  \\
 & Adaptation & 1.02 (2.07) & 2.18 (1.81) & 2.41 (1.25) \\ \hline
\multirow{2}{*}{Cognitive} & None & 3.34 (3.25) & 10.53 (3.85) & 15.56 (3.66)  \\
 & Adaptation & 3.31 (3.37) & 9.55 (3.74) & 12.77 (4.47) \\ \hline
\multirow{2}{*}{Physical} & None & 1.50 (1.82) & 3.31 (3.87) & 1.70 (2.05)  \\
 & Adaptation & 1.59 (2.05) & 3.01 (3.69) & 2.12 (2.90) \\ \hline
\multirow{2}{*}{Speech} & None & 0.06 (0.24) & 0.79 (0.68) & 2.03 (0.88)  \\
 & Adaptation & 0.51 (1.18) & 0.65 (1.3) & 0.83 (1.41) \\ \hline
\multirow{2}{*}{Overall} & None & 9.77 (5.17) & 28.65 (6.03) & 44.04 (5.07)  \\
 & Adaptation & 10.55 (5.69) & 27.05 (6.42) & 39.76 (6.65) \\ \hline

\end{tabular}

\end{table}

The tracking task performance was calculated using the RMSE in pixels between the center of the object and the center of the cross hairs (Figure \ref{fig:adapted_matb} (a)). The resulting descriptive statistics are provided in Table \ref{table:tracking}. The tracking task was automated during the underload condition, when not using the adaptive system and during the \textit{Interaction} adaptation type; thus, no corresponding results are presented. The \textit{No Adaptation} condition produced the lowest performance. The highest performance was achieved using the \textit{Both} adaptation condition for the underload and normal load conditions, while the \textit{Interaction} adaptation condition achieved the highest performance for the overload condition. This overload condition result is attributed to the tracking task being automated in the \textit{Both} and \textit{Autonomy} conditions. The higher tracking errors occurred when the participants were completing the communications task prior to the system identifying an overloaded workload state and automating the tracking task. Participants were able to complete the tracking task and the communications task simultaneously in the \textit{Interaction} condition, due to being able to verbally interact with the communication task. 
 
\begin{table}[h]
     \centering
     \caption{\label{table:tracking}Tracking Task: RMSE Performance Means (Std. Dev.) by Evaluation Type. \textbf{Note:} Lower is Better.}
     \begin{tabular}{l c c c}
         \textbf{Adaptation Type} & \textbf{Underload} & \textbf{Normal Load} & \textbf{Overload} \\ \hline
          None &  - & 140.59 (93.83) & 200.28 (111.6) \\
          Both          & \textbf{84.87 (55.83}) & \textbf{87.50 (52.44}) & 126.14 (86.37) \\
          Autonomy      & 100.82 (67.09) & 119.62 (74.47) & 115.84 (75.45) \\
          Interaction   & - & 89.57 (59.10) & \textbf{112.98 (70.83)} \\ \hline
          %Adaption    &  88.64 (58.53) & 99.20 (64.66) & 111.50 (74.76) \\ \hline
     \end{tabular}
 \end{table}

The participants were required to maintain the resource management task's primary fuel tanks' levels. The overall percentage of time the tanks were in range by adaptation type are provided in Table \ref{tab:tir}. The participants maintained the fuel levels the best when the system adapted \textit{Interactions} for each workload condition. The participants in the \textit{No Adaptation} condition achieved the lowest performance for the underload and normal load conditions, while the \textit{Both} adaptation type performed the worst during the overload condition.
 
 \begin{table}[h]
     \centering
     \caption{\label{tab:tir} Resource Management Task: Time in Range (\%) by Evaluation Type. \textbf{Note:} Higher is Better.}
     \begin{tabular}{l c c c}
         \textbf{Adaptation Type} & \textbf{Underload} & \textbf{Normal Load} & \textbf{Overload} \\ \hline
          None &  79 & 79 & 68 \\
          Both & 84 & 76 & 61 \\
          Autonomy & 91 & 85 & 93 \\
          Interaction & \textbf{92} & \textbf{91} & \textbf{99} \\ \hline
          %Adaption    &  84 & 76 & 61 \\ \hline
     \end{tabular}
 \end{table}

The system monitoring task consisted of resetting lights and gauges, when they went out of range. The participants' descriptive statistics for reaction time to the out of range lights or gauges by adaptation type and workload condition are provided in Table \ref{tab:sys_rt}. The participants achieved the best performance in the underload and overload conditions when tasks were automated and the best performance in the normal load condition when interactions were adapted. The none condition resulted in the worst overall reaction times. 
 
 \begin{table}[h]
 \vspace{0.1cm}
     \centering
     \caption{\label{tab:sys_rt} System Monitoring Reaction Time Means (Std. Dev.) by Evaluation Type. \textbf{Note:} Lower is Better.}
     \begin{tabular}{l c c c}
         \textbf{Adaptation Type} & \textbf{Underload} & \textbf{Normal Load} & \textbf{Overload} \\ \hline
          None &  4.32 (4.16) & 5.38 (3.52) & 6.25 (4.30) \\
          Both & 3.14 (2.19) & 4.53 (2.55) & 5.68 (3.67) \\
          Autonomy & \textbf{2.86 (1.15)} & 5.63 (3.45) & \textbf{4.52 (3.18)} \\
          Interaction & 5.68 (3.67) & \textbf{4.13 (2.60)} & 5.83 (3.77) \\ \hline
          %Adaption    &  3.02 (1.74) & 4.66 (2.86) & 5.40 (3.61) \\ \hline
     \end{tabular}
 \end{table}
 
A failure occurred if a light or gauge was not corrected within fifteen seconds of when the light or gauge went out of range. The system monitoring success rate by adaptation type is provided in Table \ref{tab:sys_sr}. The participants were the least successful when no adaptation occurred. The \textit{Both} and \textit{Autonomy} adaptation conditions achieved roughly the same performance in the underload and normal load conditions, but the best performance under the \textit{Autonomy} adaptation condition for the overload condition. The highest performance was achieved in the \textit{Interaction} condition for the normal load condition.
 
\begin{table}[h]
     \centering
     \caption{\label{tab:sys_sr}System Monitoring Success Rate (\%) by Adaptation Type and Workload Condition. \textbf{Note:} Higher is Better.}
     \begin{tabular}{l c c c}
          \textbf{Adaptation Type} & \textbf{Underload} & \textbf{Normal Load} & \textbf{Overload} \\ \hline
          None &  69 & 79 & 60 \\
          Both & \textbf{100} & 95 & 75 \\
          Autonomy & \textbf{100} & 94 & \textbf{85} \\
          Interaction & 92 & \textbf{99} & 76 \\ \hline
          %Adaption    &  97.91 & 95.98 & 77.71 \\ \hline
     \end{tabular}
     \vspace{-0.25cm}
 \end{table}
 
The participants responded to simulated air-traffic control messages during the communication task. The average (Std. Dev.) time it took for participants to respond to these messages by adaptation condition is provided in Table \ref{tab:comm_rt}. The participants responded to messages quicker with the \textit{Both} adaptation condition and were the slowest when \textit{no adaptation} occurred. Similar reaction times were expected when no adaptation occurred and during the \textit{Interaction} adaptation condition for the normal load condition, as the speech interaction modality adaptation never occurred during the normal load condition.

 \begin{table}[!h]
     \centering
     \caption{\label{tab:comm_rt}Communications Reaction Time Means (Std. Dev.) by Evaluation Type. \textbf{Note:} Lower is Better.}
     \begin{tabular}{l c c c}
         \textbf{Adaptation Type} & \textbf{Normal Load} & \textbf{Overload} \\ \hline
          None & 10.41 (1.79) & 9.83 (4.36) \\ 
          Both &  \textbf{8.68 (4.57)} & \textbf{3.46 (4.07)} \\
          Autonomy &  9.17 (4.9) & 5.12 (5.49) \\
          Interaction & 10.35 (3.36) & 4.36 (5.05) \\ \hline
     \end{tabular}
     \vspace{-0.25cm}
 \end{table}

The overall task performance calculations are provided in Table \ref{tab:between_overall_perf}. Directly comparing the overall task performance values between the adaptive conditions is confounded by what tasks were active during the conditions, as the overall task performance value was based on the active task set. For example, the tracking task was inactive during the underload condition for the \textit{Interaction} adaptation type, but was active for the \textit{Both} and \textit{Autonomy} types. Thus, the tracking task performance may deflate the overall performance value artificially. Some general trends can be extrapolated. Adapting interaction modalities for each workload condition resulted in higher task performance than the \textit{No Adaptation} condition. This comparison is not confounded, as both conditions had the same active task set. 

\begin{table}[!h]
%\vspace{-0.2cm}
     \centering
     \caption{\label{tab:between_overall_perf} Calculated Overall Performance Descriptive Statistics by Adaptation Type and Workload Condition. \textbf{Note:} Higher is Better.}
     \begin{tabular}{l c c c}
          \textbf{Adaptation Type} & \textbf{Underload} & \textbf{Normal Load} & \textbf{Overload} \\ \hline
          None & 0.85 (0.12) & 0.72 (0.20) & 0.56 (0.16)  \\
          Both & 0.81 (0.12) & \textbf{0.83 (0.13}) & 0.66 (0.27)\\
          Autonomy & 0.82 (0.06) & \textbf{ 0.81 (0.14}) & \textbf{0.77 (0.16)} \\
          Interaction & \textbf{0.98 (0.04)} & \textbf{0.83 (0.12)} & 0.72 (0.09) \\ \hline
          %Adaption    &  97.91 & 95.98 & 77.71 \\ \hline
     \end{tabular}
    \vspace{-0.25cm}
 \end{table}

The subjective results consisted of the In-Situ ratings and the NASA-TLX. The associated descriptive statistics for only the overall ratings are provided in Table \ref{tab:subjective}. The overall In-Situ ratings (theoretical range 6-30) demonstrate that participants perceived lower workload levels (Mann-Whitney $U=207.5$, $p<0.01$) during the overload conditions when using SAHRTA vs. not using SAHRTA. There was no perceived workload difference during the underload and normal load conditions, as determined by the Mann-Whitney U test. There was no significant difference between the overall NASA-TLX ratings administered after the trial. These results were to be expected, due to the relatively small number of participants, the ratings' subjective nature, and the between-subjects experimental design.  

\begin{table}[!h]
\vspace{-0.3cm}
\centering
\caption{\label{tab:subjective} Descriptive Statistics for the Overall In-Situ Workload Ratings and the NASA-TLX.}

\begin{tabular}{c |c c c| c} \hline
 & \multicolumn{3}{c|}{\textbf{Overall In-Situ Ratings}} & \\
 \textbf{Adaptation} & \textbf{UL} & \textbf{NL} & \textbf{OL} & \textbf{TLX} \\ \hline
None & 8.6 (2.5) & 14.1 (3.7) & 18.7 (3.5) & 60.0 (12.3)  \\
Adaptation & 9.1 (3.1) & 14.1 (2.7) & 15.2 (4.1) & 60.2 (14.0) \\ \hline
%Mann-Whitney & 144.5 & 157.5 & 207.5** & 39.0\\ \hline  
% None & 8.60 (2.54) & 14.10 (3.68) & 18.70 (3.53) & 60.00 (12.25)  \\
% Adaptation & 9.12 (3.07) & 14.06 (2.74) & 15.17 (4.06) & 60.17 (13.97) \\ \hline

\end{tabular}
\vspace{-0.5cm}
\end{table}

\section{Discussion}

SAHRTA targets adaptations to specific workload channels in order to augment task performance. Hypothesis $\mathbf{H^A_1}$ predicted that participants will experience lower workload in the overload condition and higher workload for the underload condition when using SAHRTA. The hypothesis was supported for the overload condition, but not for the underload condition. Failing to support the hypothesis for the underload condition was attributed to the SAHRTA's sub-optimal approach of transitioning the tracking task out of automatic mode. An adaptive system may allocate other tasks to the participant, but such an approach is not feasible with the NASA MATB-II. Adaptive systems cannot recreate communication requests to which the participant can respond or make the system monitoring task's alarms go out of range, as the alarms represented system states. 

Targeting adaptations to overall workload and its contributing components was expected to increase task performance. Hypothesis $\mathbf{H^A_2}$ predicted that higher task performance will be achieved using SAHRTA. This hypothesis was supported for the between adaptations analysis, as the highest task performance for each NASA MATB-II task occurred in one of the adaptive conditions (i.e., Both, Autonomy, or Interaction). %Fully supporting this hypothesis does not provide insight into which adaptations were most effective for each NASA MATB-II task; thus, additional analysis were conducted to investigate the effectiveness of automating tasks and adapting interactions.

The system adaptions (autonomy levels and interactions) were analyzed by examining their impact on workload and task performance. Automating tasks did not have the expected impact on workload, but did allow the participants to focus on the non-automated tasks and generally achieve higher performance. However, similar performance was achieved when adapting either the interactions or the autonomy. This result was attributed to participants being able to complete the tracking, system monitoring, and communications task simultaneously when the speech interaction modality was active for the communications task. This result demonstrates that the adaptive system was able to balance workload across the workload channels. Specifically, the physical workload requirement of walking between the tasks was allocated to the participant's speech workload channel.

Workload was used to determine SARTA's adaptations in order to augment task performance, as workload is an indirect measure of task performance. Measuring task performance directly may appear as a more accurate and informative measure on which to base system adaptations, but there are several limitations to such an approach. First, task performance may be difficult to measure directly in dynamic task environments (i.e., first response domains); thus, relying solely on task performance limits the range of environments in which the adaptive system can be deployed. Second, an overall task performance measure does not provide meaningful information about how a specific interaction will affect the human (e.g., will an auditory modality decrease task performance due to resource conflicts). These limitations demonstrate that relying on workload as a surrogate for task performance provides for more robust system adaptations and allows SAHRTA to be deployed in multiple task environments. 

%The current adaptive system considers the workload component information as independent estimates, but these estimates are likely correlated. A task analysis may be used to identify the common means to complete the subtasks, which will identify what specific workload channels will be used. Additionally, a correlation or principle component analysis between the component estimates may help determine what the interdependicies are between the components. This information may allow for a more representative overall workload estimate that weights each workload component accordingly. Having a more representative overall estimate will allow the adaptive system to better reason how an adaptation will impact the human. For example, using a visual modality (i.e., text) for an alarm will have a cognitive component associated with the alarm. If the human is nearing an overloaded state, the system may reason that the visual modality will overload the human's cognitive workload and choose not to use that specific modality, as an auditory modality may be more appropriate.

SAHRTA depended on continuous workload estimates, which permits treating the system as a multi-variate control system. The workload estimates act as sensors with the same update rates (5-seconds), while the system adaptations can be considered to be system corrections. The update rate of the system corrections can impact the control system's stability and controllability. If these corrections are updated too frequently, the system will oscillate, resulting in unstable system states. For example, invoking and revoking autonomy continuously may increase the human's workload level, as the human has to reassess the task states constantly in order to have appropriate system awareness. Adapting too slowly will result in the system never reaching the desired steady-state (i.e., the human is performing optimally). For example, if autonomy decisions are considered once every 5-minutes, then the human may be in an overloaded state for at most 5-minutes. Additionally, relying on data from 5-minutes prior is insufficient for determining system interactions, such as choosing an auditory alarm for the system monitoring task. Choosing an appropriate adaptation update rate is not trivial and is likely domain and task specific. Adaptive system designers need real-world data from the specific domain and tasks in order to determine proper adaptation update rates and their corresponding impact on the adaptive system's stability.

Future work will extend SAHRTA to different domains (e.g., peer-based human-teams). Such domains tend to be more dynamic, thus; activity recognition will be used to determine the human's current task focus. Additionally, the autonomy level adaptations will occur for neglected tasks and the human's current task in order to better mitigate the overload condition. 

%Other aspects of control theory may be applicable to the adaptive system. A proportional integral derivative controller may be applied in order to maintain the human's performance at a desired level. The controller can provide information regarding how much to adapt. If the human is overloaded, then the controller can determine how much autonomy is needed in order to normalize the human's workload level. However, the adaptive system needs to have the necessary level of controllability in order to invoke the necessary level of autonomy. For example, the current adaptive system has two autonomy levels (on/off) for each task and automating neglected tasks did not impact the human's workload level effectively. The controller may increase the autonomy of each task, which may reduce vigiliance. Thus, the current system's controllability may be considered to be low. Implementing more than two levels of autonomy for each task will increase the system's controllability and will allow for a proportional integral derivative controller to be more effective.

\section{Conclusion}

SAHRTA was demonstrated to improve task performance over a version of the system without adaptations. This task performance increase was primarily attributed to the system being able to select the appropriate interaction modality for the system monitoring and communications task. Automating neglected tasks was beneficial, as the automation allowed participants to better focus their attention and increase overall task performance. Further analyses and evaluations are needed to better understand how to adapt system autonomy levels and interactions appropriately, but SAHRTA is a necessary step towards effective human-robot teaming architectures.

\bibliographystyle{IEEEtran}
\bibliography{dissertation_papers}

\end{document}